\PassOptionsToPackage{unicode}{hyperref}
\PassOptionsToPackage{hyphens}{url}
\PassOptionsToPackage{dvipsnames,svgnames,x11names}{xcolor}
\documentclass[
  12pt,
  a4paper,
]{article}
\usepackage{xcolor}
\usepackage[margin=3cm]{geometry}
\usepackage{amsmath,amssymb}
\usepackage{iftex}
\ifPDFTeX
  \usepackage[T1]{fontenc}
  \usepackage[utf8]{inputenc}
  \usepackage{textcomp} 
\else 
  \usepackage{unicode-math} 
  \defaultfontfeatures{Scale=MatchLowercase}
  \defaultfontfeatures[\rmfamily]{Ligatures=TeX,Scale=1}
\fi
\usepackage{lmodern}
\ifPDFTeX\else
\fi
\IfFileExists{upquote.sty}{\usepackage{upquote}}{}
\IfFileExists{microtype.sty}{
  \usepackage[]{microtype}
  \UseMicrotypeSet[protrusion]{basicmath} 
}{}
\makeatletter
\@ifundefined{KOMAClassName}{
  \IfFileExists{parskip.sty}{%
    \usepackage{parskip}
  }{
    \setlength{\parindent}{0pt}
    \setlength{\parskip}{6pt plus 2pt minus 1pt}}
}{
  \KOMAoptions{parskip=half}}
\makeatother
\usepackage{longtable,booktabs,array}
\usepackage{calc} 
\usepackage{etoolbox}
\makeatletter
\patchcmd\longtable{\par}{\if@noskipsec\mbox{}\fi\par}{}{}
\makeatother
\IfFileExists{footnotehyper.sty}{\usepackage{footnotehyper}}{\usepackage{footnote}}
\makesavenoteenv{longtable}
\usepackage{graphicx}
\makeatletter
\newsavebox\pandoc@box
\newcommand*\pandocbounded[1]{
  \sbox\pandoc@box{#1}%
  \Gscale@div\@tempa{\textheight}{\dimexpr\ht\pandoc@box+\dp\pandoc@box\relax}%
  \Gscale@div\@tempb{\linewidth}{\wd\pandoc@box}%
  \ifdim\@tempb\p@<\@tempa\p@\let\@tempa\@tempb\fi
  \ifdim\@tempa\p@<\p@\scalebox{\@tempa}{\usebox\pandoc@box}%
  \else\usebox{\pandoc@box}%
  \fi%
}
\def\fps@figure{htbp}
\makeatother
\usepackage{xparse}
\NewDocumentCommand\citeproctext{}{}

\makeatletter
 \let\@cite@ofmt\@firstofone
 \def\@biblabel#1{}
 \def\@cite#1#2{{#1\if@tempswa , #2\fi}}
\makeatother
\newlength{\cslhangindent}
\newlength{\csllabelwidth}
\newenvironment{CSLReferences}[2] 
 {\begin{list}{}{%
  \setlength{\itemindent}{0pt}
  \setlength{\leftmargin}{0pt}
  \setlength{\parsep}{0pt}
  \ifodd #1
   \setlength{\leftmargin}{\cslhangindent}
   \setlength{\itemindent}{-1\cslhangindent}
  \fi
  \setlength{\itemsep}{#2\baselineskip}}}
 {\end{list}}
\usepackage{calc}

\providecommand{\tightlist}{%
  \setlength{\itemsep}{0pt}\setlength{\parskip}{0pt}}
\usepackage{graphicx}
\usepackage{color}
\makeatletter
\@ifpackageloaded{subfig}{}{\usepackage{subfig}}
\@ifpackageloaded{caption}{}{\usepackage{caption}}
\AtBeginDocument{%

}
\AtBeginDocument{%

}
\newcounter{pandoccrossref@subfigures@footnote@counter}
{\end{figure}%
\addtocounter{footnote}{-\value{pandoccrossref@subfigures@footnote@counter}}
\@for\f:=\global@pandoccrossref@subfigures@footnotes\do{\stepcounter{footnote}\footnotetext{\f}}%
\gdef\global@pandoccrossref@subfigures@footnotes{}}
\@ifpackageloaded{float}{}{\usepackage{float}}
\floatstyle{ruled}
\@ifundefined{c@chapter}{\newfloat{codelisting}{h}{lop}}{\newfloat{codelisting}{h}{lop}[chapter]}
\floatname{codelisting}{Listing}

\makeatother
\usepackage{bookmark}
\IfFileExists{xurl.sty}{\usepackage{xurl}}{} 
\hypersetup{
  pdftitle={RAG with Differential Privacy},
  pdfauthor={Nicolas Grislain},
  colorlinks=true,
  linkcolor={blue},
  filecolor={Maroon},
  citecolor={red},
  urlcolor={magenta},
  pdfcreator={LaTeX via pandoc}}

\title{RAG with Differential Privacy}
\author{Nicolas Grislain}
\date{December 5, 2024}

\begin{document}
\maketitle
\begin{abstract}
Retrieval-Augmented Generation (RAG) has emerged as the dominant
technique to provide \emph{Large Language Models} (LLM) with fresh and
relevant context, mitigating the risk of hallucinations and improving
the overall quality of responses in environments with large and fast
moving knowledge bases. However, the integration of external documents
into the generation process raises significant privacy concerns. Indeed,
when added to a prompt, it is not possible to guarantee a response will
not inadvertently expose confidential data, leading to potential
breaches of privacy and ethical dilemmas. This paper explores a
practical solution to this problem suitable to general knowledge
extraction from personal data. It shows \emph{differentially private
token generation} is a viable approach to private RAG.
\end{abstract}

\section{Introduction}\label{introduction}

Retrieval-Augmented Generation (RAG, (Lewis et al. 2021)) has become a
popular approach to enhance the capabilities of Large Language Models
(LLMs) by supplying them with up-to-date and pertinent information. This
method is particularly valuable in environments where knowledge bases
are large and rapidly evolving, such as news websites, social media
platforms, or scientific research databases. By integrating fresh
context, RAG helps mitigate the risk of ``hallucinations''---instances
where the model generates plausible but factually incorrect
information---and significantly improves the overall quality and
relevance of the responses generated by the LLM.

However, incorporating external documents into the generation process
introduces substantial privacy concerns. When these documents are
included in the input prompt for the LLM, there is no foolproof way to
ensure that the generated response will not accidentally reveal
sensitive or confidential data (Qi et al. 2024). This potential for
inadvertent data exposure can lead to serious breaches of privacy and
presents significant ethical challenges. For instance, if an LLM is used
in a healthcare setting and it accidentally includes patient information
from an external document in its response, it could violate patient
confidentiality and legal regulations.

This paper describes a practical solution (DP-RAG) aimed at addressing
these privacy concerns with \emph{Differential Privacy} (DP). The
solution is based on two pillars:

\begin{itemize}
\tightlist
\item
  A method to collect documents related to the question in a way that
  does not prevent its output to be used in a DP mechanism.
\item
  A method to use the collected documents to prompt a LLM and produce a
  response with DP guarantees.
\end{itemize}

The paper describes also some empirical tests and shows that
\emph{DP-RAG} is most effective in context where enough documents give
elements of response.

\section{Related Work}\label{related-work}

In general there are two families of approaches to add new knowledge to
an LLM. The first is \emph{Fine Tunning} (FT) and the other is
\emph{Retrieval Augmented Generation} (RAG). In both these approaches,
adding privacy can be done, through simple heuristics with human
validation such as \emph{masking} or using a systematic and
principle-based approach such as \emph{Differential Privacy}.

\subsection{Private Fine-Tuning}\label{private-fine-tuning}

A straightforward approach to adding knowledge to an existing LLM is to
continue its training with the new knowledge, to Fine Tune (FT) it.
However, this raises challenges when dealing with private data, as LLMs
tend to memorize training data. (see (Shokri et al. 2017) or (Carlini et
al. 2021)).

To mitigate this privacy risk, it is possible to redact sensitive
content prior to the FT process (aka. \emph{masking}), but this
operation is not very reliable and requires judgment on what should be
redacted. This is a difficult manual operation based on the perceived
sensitivity of each field and how it can be used to re-identify an
individual, especially when combined with other publicly available data.
Overall, it is very easy to get wrong; leaning too much on the side of
prudence can yield useless data, while trying to optimize utility may
result in leaking sensitive information.

A solution to this problem is to leverage \emph{Differential Privacy}, a
theoretical framework enabling the computation of aggregates with formal
privacy garantees (See (Dwork, Roth, et al. 2014)).

The most common approache to Private LLM FT is to use
Differentially-Private-Stochastic-Gradient-Descent (DP-SGD, see (Abadi
et al. 2016) and (Ponomareva et al. 2023)). DP-SGD is about clipping
gradients and adding them some noise while running your ordinary SGD (or
standard variants such as \emph{Adam}, etc.). This method requires the
data to be organized per \emph{privacy unit} (typically a privacy unit
will be a user). Every training example should belong to one and only
one privacy unit\footnote{Note that observations (examples) can be
  grouped into composite observations if one user contributes to many
  observations.}.

But, when new documents are frequently added to the private knowledge
base FT may not be the best approach.

\subsection{Private RAG}\label{private-rag}

When FT is not the best approach to adding new knowledge and RAG would
be preferred, DP-FT cannot help with privacy. In these cases, DP can
still be leveraged in different ways. A straightforward approach to DP
RAG is to generate synthetic documents with differential privacy out of
the private knowledge base and then retrieve documents from this
synthetic knowledge base instead of the private one. Another approach is
to generate the LLM response in a DP way.

The approach of generating synthetic documents usable for RAG in
privacy-sensitive contexts has been explored by (Zeng et al. 2024) but
without DP guarantees. There are three main approaches to the problem of
generating DP Synthetic Data (SD):

\begin{itemize}
\tightlist
\item
  Fine-Tuning a pre-trained generative model with DP to generate
  synthetic documents.
\item
  Use some form of automated prompt tuning to generate synthetic prompts
  or context documents.
\item
  And use DP aggregated generation.
\end{itemize}

Fine-Tuning a pre-trained generative model with DP can be done with
DP-SGD ((Abadi et al. 2016) and (Ponomareva et al. 2023)) as mentioned
above. An application to synthetic text generation is described there:
(Yue et al. 2023). This method is technically complex, as, DP-SGD can be
challenging to implement efficiently (Bu et al. 2023).

In (Hong et al. 2024), the authors use an automated prompt tuning
technique developed in (Sordoni et al. 2023) and (Zhou et al. 2023) and
make it differentially private. From the evaluations presented, it seems
to compare favorably to DP-FT synthetic data approaches. Similar
methods, based on DP-automated prompt tuning are exposed in (Lin et al.
2024) for images and (Xie et al. 2024) for text.

A last approach to generating synthetic data is based on DP aggregation
of data.(Lebensold et al. 2024) or (Wu et al. 2023) show how to
aggregate images or text in their embedding space (aka. Embedding Space
Aggregation). Aggregating data privately is also the approach of (Tang
et al. 2024), but they do it at the token level.

This last method greatly inspired the approach described in this
document, though not for SD, but to directly generate RAG output from
private documents.

\section{DP-RAG}\label{dp-rag}

To overcome the limitations of DP FT or SD-based RAG, we developed and
tested DP-RAG: a novel approach, build upon recent works on DP
In-Context Learning (ICL) such as (Wu et al. 2023) and particularly
(Tang et al. 2024).

\begin{itemize}
\tightlist
\item
  Contrary to (Wu et al. 2023), we aggregate outputs token by token.
\item
  Our token aggregation method is different from both methods exposed
  in: (Tang et al. 2024) (\emph{Gaussian} and \emph{Report Noisy Max}).
\item
  Because we implement the full RAG system, we developed a method to
  collect the \emph{top-k} most similar documents in a way that does not
  jeopardize the possibility to run a DP mechanism on them.
\end{itemize}

\subsection{Overview of DP-RAG}\label{overview-of-dp-rag}

DP-RAG is made of two main components:

\begin{itemize}
\tightlist
\item
  A method to collect documents related to the question in a way that
  does not prevent its output to be used in a DP mechanism.
\item
  A method to use the collected documents to prompt a LLM and produce a
  response with DP guarantees.
\end{itemize}

\begin{figure}
\centering
\includegraphics[width=100mm,height=\textheight,keepaspectratio]{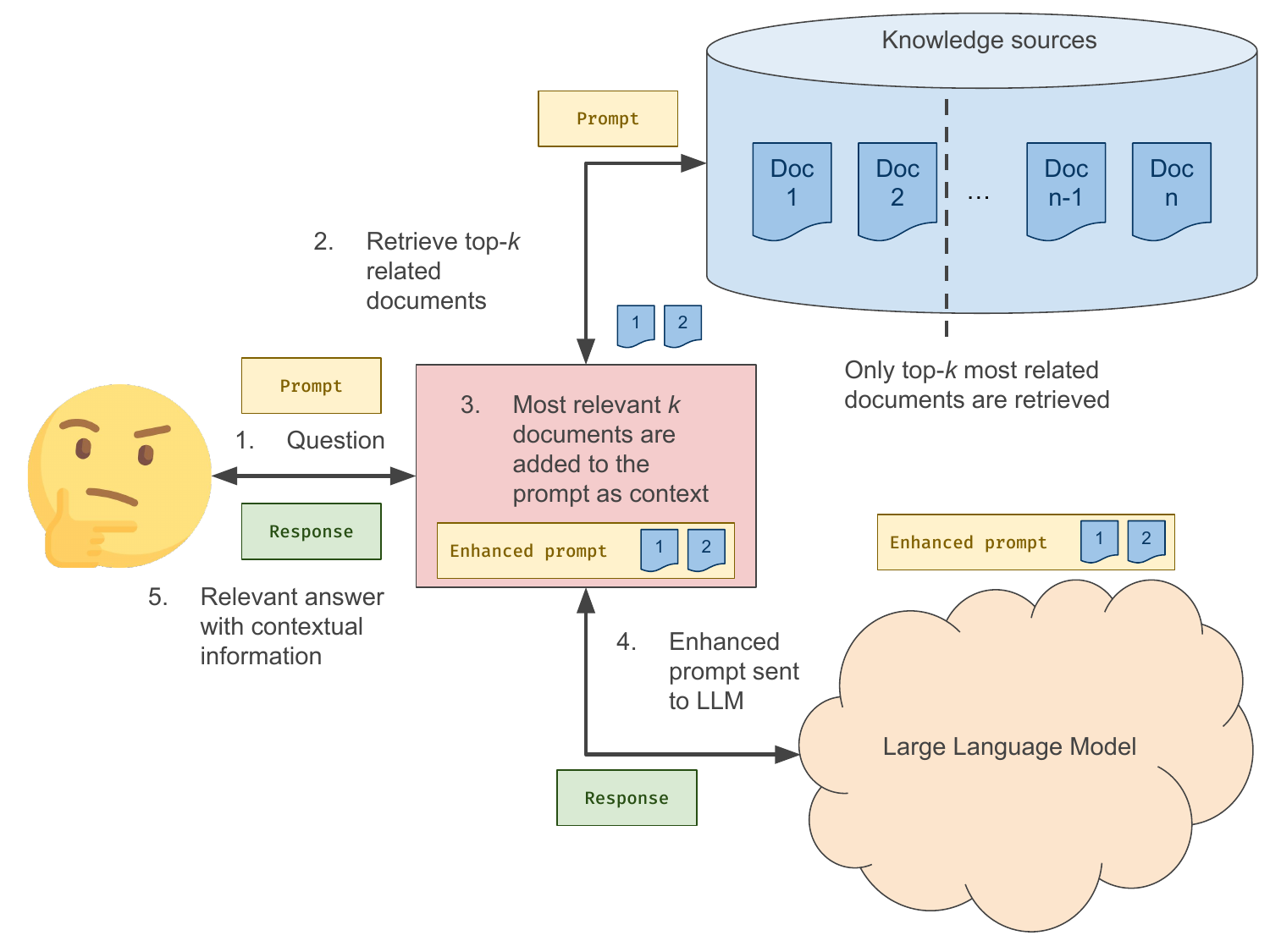}
\caption{A broad picture of how RAG works}\label{fig:rag}
\end{figure}

To understand the need for these components, let's describe what RAG is
usually made of (see also (Lewis et al. 2021)) and introduce some
notations (see Fig.~\ref{fig:rag}).

A LLM: \(\mathcal{L}\) is a function, taking some text, in the form of a
sequence of tokens: \(x = \left<x_1, x_2, \ldots, x_n\right>\) as input
and outputting a probability distribution of the next token \(x_{n+1}\)
conditional on \(x\):
\[\mathcal{L}(s, x) = \mathcal{L}(s, \left<x_1, x_2, \ldots, x_n\right>) = \Pr(x_{n+1} = s | \mathcal{L}, x_1, x_2, \ldots, x_n)\]

We assume we have a set of \(N\) documents:
\(D = \left\{d_1, d_2, \ldots, d_N\right\} \subset \mathcal{D}\)
containing domain specific knowledge. These documents are also sequences
of tokens: \(d_i = \left<d_{i,1}, d_{i,2}, \ldots, d_{i,l_i}\right>\).
We will, for simplicity, denote \(\left<d_i, d_j\right>\) the
concatenation of two sequences of token, or a sequence and one token.

We also assume we have a similarity function
\(S: \mathcal{D}^2 \mapsto [-1, 1]\) which value is close to 1 when two
documents are very similar, close to 0 when independent, and close to -1
when conveying opposite meaning. In this work \(S\) will be the
\emph{cosine similarity} between some embeddings of the documents,
mapping them to some adequate \(d\)-dimensional vector space:
\(\mathbb{R}^d\):
\[S(d_i, d_j) = \frac{\left<E(d_i), E(d_j)\right>}{\|E(d_i)\|_2\|E(d_j)\|_2}\]

When receiving a query in the form of a sequence of token:
\(q = \left<q_1, q_2, \ldots, q_{n_q}\right>\), the similarity between
\(q\) and each document is computed and the top \(k\) documents in term
of similarity are collected:
\[d_{i_1}, d_{i_2}, \ldots d_{i_k} \text{ with } S(q, d_{i_1}) \geq S(q, d_{i_2}) \geq \ldots \geq S(q, d_{i_N})\]

Then a new query \(q_{RAG}\) is built by concatenating the original
query \(q\) with the top \(k\) documents and other elements (the
operation is denoted \(\left<\cdot, \ldots ,\cdot\right>_{RAG}\))
\[q_{RAG} = \left<q, d_{i_1}, d_{i_2}, \ldots d_{i_k}\right>_{RAG}\]

The augmented query is then sent to the LLM to compute the distribution
of the next token (the first token of the response)
\[\mathcal{L}\left(r_1, \left<q, d_{i_1}, d_{i_2}, \ldots d_{i_k}\right>_{RAG}\right)\]

The token is generated by sampling according to the
distribution\footnote{or proportionally to some power \(1/T\) of the
  distribution} or by selecting the mode of the distribution\footnote{the
  most likely token or the limit when \(T\) goes to \(0\)}.

The tokens of the response are then generated one by one in an
auto-regressive manner. The generated response tokens are concatenated
to the input sequence:
\[\mathcal{L}\left(r_{j+1}, \left<\left<q, d_{i_1}, d_{i_2}, \ldots d_{i_k}\right>_{RAG}, r_1, r_2,\ldots, r_j\right>\right)\]

In the private variant of the problem (DP-RAG), we also assume the
documents are \emph{privacy sensitive}, and make the additional
assumption that each document relates to only one individual that we
call \emph{privacy unit} (PU)\footnote{Such structuring of documents by
  privacy unit can sometime be achieved by cutting documents and
  grouping all the content relative to one PU in one document.}.

\subsection{Differential Privacy and its application to
RAG}\label{differential-privacy-and-its-application-to-rag}

A (randomized) algorithm: \(\mathcal {A}\) provides
\((\epsilon,\delta)\)-Differential Privacy \emph{if and only if} for all
event \(S\) and neighboring datasets \(D_0\) and \(D_1\), we have:
\[\Pr[{\mathcal {A}}(D_{0})\in S]\leq e^{\varepsilon }\Pr[{\mathcal {A}}(D_{1})\in S]+\delta\]

This means that for datasets that differ by one individual
(i.e.~neighboring datasets) the algorithm's outputs are statistically
indistinguishable. This property guarantees that no bit of information
can reasonably be learned about an individual. See (Dwork, Roth, et al.
2014) for a thorough introduction to DP.

\begin{figure}
\centering
\includegraphics[width=100mm,height=\textheight,keepaspectratio]{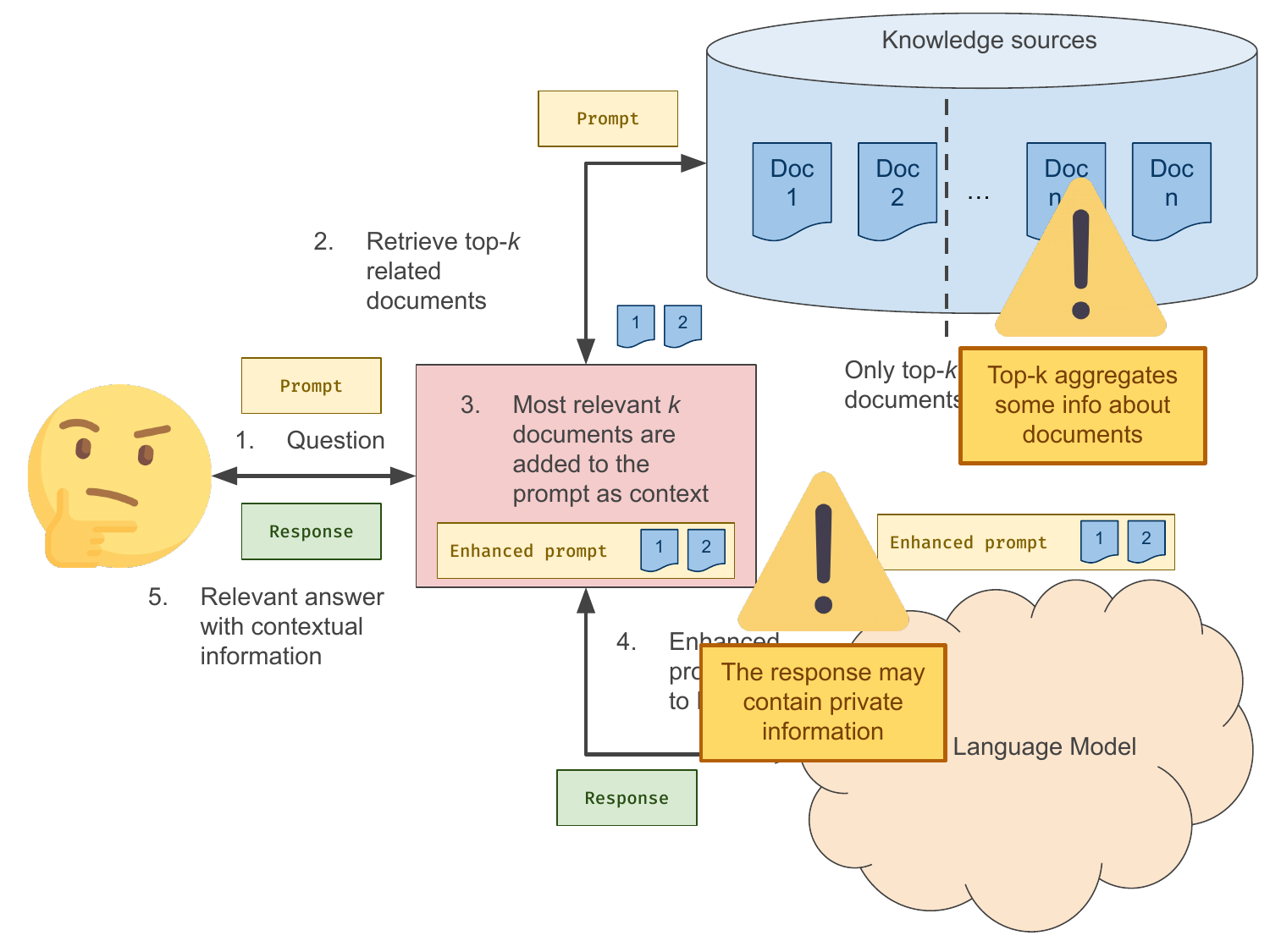}
\caption{A broad picture of the main problems to overcome when
considering DP RAG}\label{fig:ragpriv}
\end{figure}

The two main challenges to implementing RAG with DP guarantees (see
Fig.~\ref{fig:ragpriv}) consist in:

\begin{itemize}
\tightlist
\item
  aggregating the knowledge from many documents with DP,
\item
  and, more subtly, selecting the most relevant documents without
  jeopardizing our ability to apply a DP mechanism downstream.
\end{itemize}

\begin{figure}
\centering
\includegraphics[width=100mm,height=\textheight,keepaspectratio]{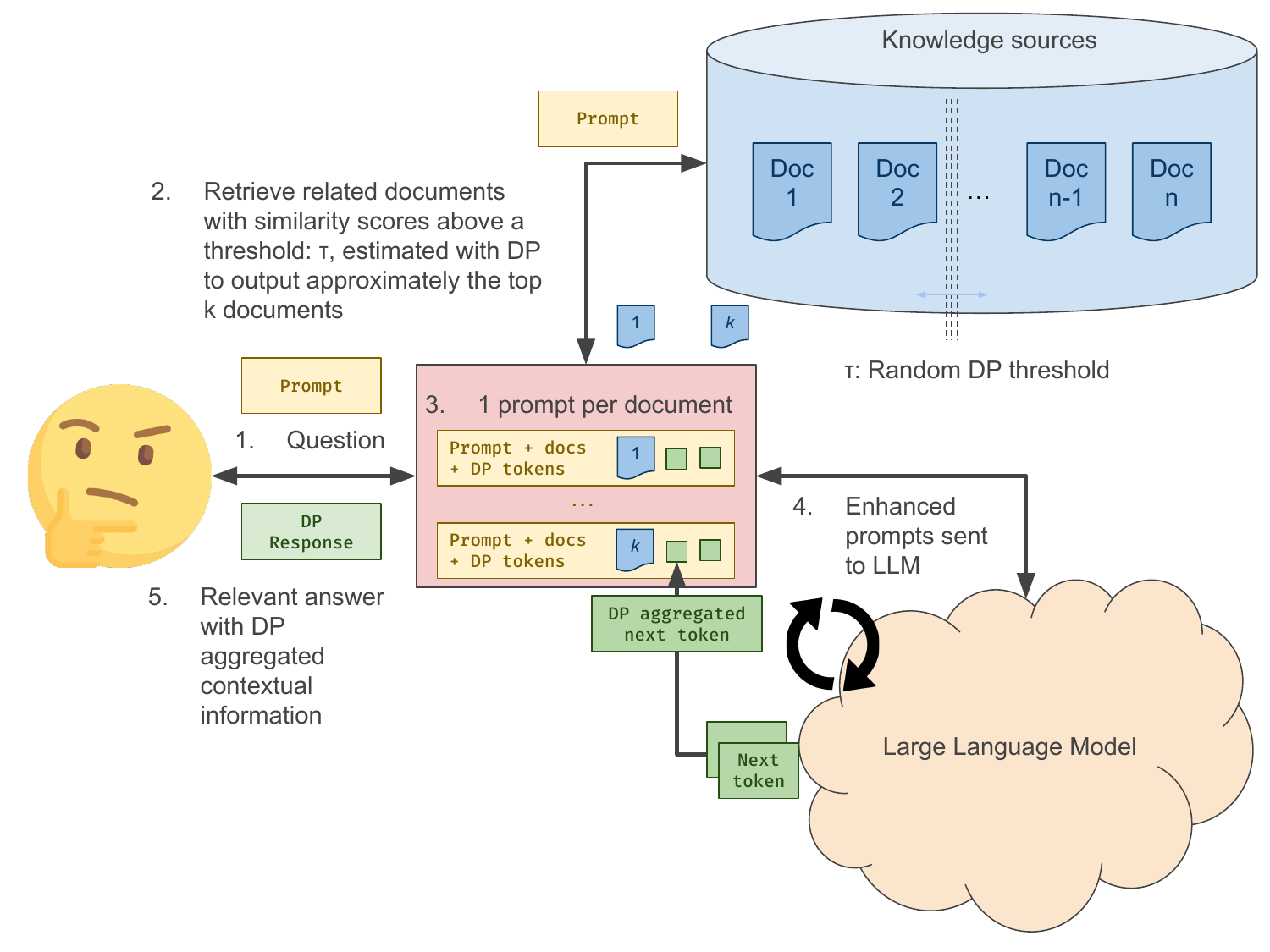}
\caption{In DP-RAG, \(k\) smaller queries are sent to the LLM, rather
than a single query (approximately) \(k\) times
larger.}\label{fig:dprag}
\end{figure}

\subsection{Privacy Unit Preserving Document
Retrieval}\label{privacy-unit-preserving-document-retrieval}

As mentioned above, DP deals with the concept of \emph{neighboring}
datasets. For this reason, it is convenient to assign each document to
one and only one individual, or \emph{privacy unity} (PU). Adding or
removing one PU, comes down to adding or removing one document. In this
context, one should be careful with the selection of the top-k most
relevant documents. Indeed, when selecting the top-k documents, adding
or removing one document may affect the selection of other documents.

In DP-RAG, the similarity of each document with the query is computed:
\[s_1, s_2,\ldots, s_N = S(q, d_1), S(q, d_2), \ldots, S(q, d_N)\]

\begin{figure}
\centering
\includegraphics[width=100mm,height=\textheight,keepaspectratio]{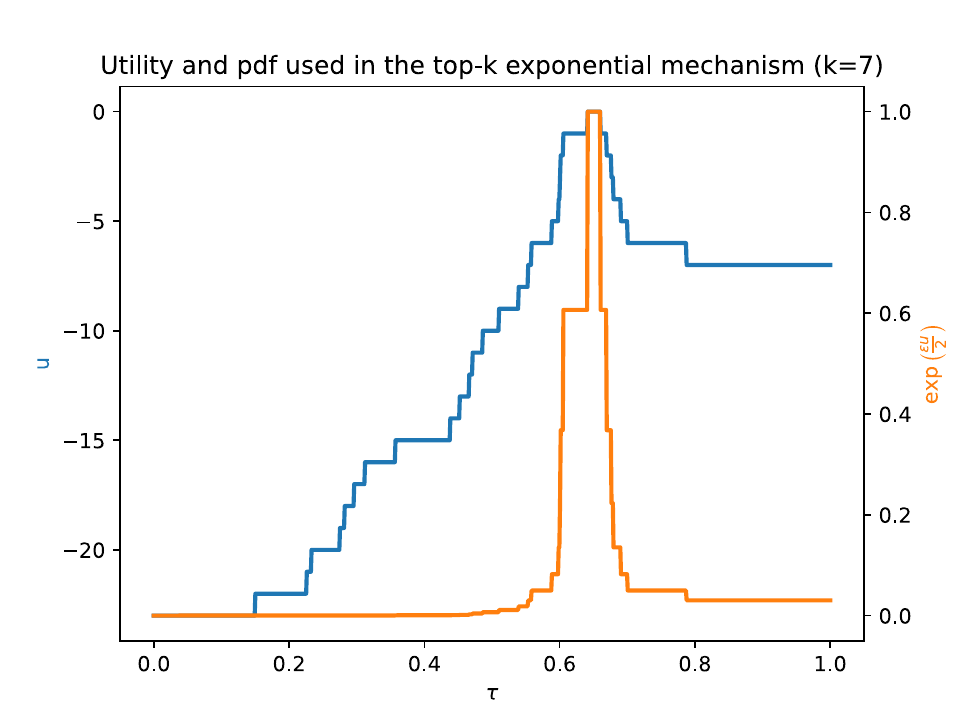}
\caption{The exponential mechanism for the top-k DP-threshold. For the
sake of clarity we chose a small number of documents: 30, and a large
\(\epsilon\): 1}\label{fig:topkexp}
\end{figure}

To estimate a threshold to select the top k documents with DP, we
designed a utility function to be plugged into an exponential mechanism
(Dwork, Roth, et al. 2014) (see Fig.~\ref{fig:topkexp}).
\[U_{top-k}(\tau): [0, 1] \mapsto \mathbb{R} = -\left|\sum_i\mathbf{1}_{[0, s_i]}(\tau)-k\right|\]

This \emph{top-k} utility has sensitivity 1, we can sample a threshold
\(\tau_{DP}\) from the probability density function:
\[\tau_{top-k}\propto\exp\left(\frac{\epsilon U_{top-k}(\tau)}{2}\right)\]

It is easy to show \(\tau_{top-k}\) is \(\epsilon\)-DP (see. (Dwork,
Roth, et al. 2014)).

The DP top-k threshold \(\tau_{top-k}\) sampled from the exponential
mechanism is then used to select all the documents whose similarity is
above \(\tau_{top-k}\).

While this threshold, works well in practice, it selects a fixed number
of documents (\textasciitilde k). We may be interested in selecting
fewer when the top scores are more concentrated on few documents (the
query is \emph{selective}), and select more when the scores are evenly
spread across many documents (the query has a low \emph{selectivity}).
To adjust to this need, we designed a slightly different utility
function:
\[U_{top-p}(\tau): [0, 1] \mapsto \mathbb{R} = -\left|\sum_i\mathbf{1}_{[0, s_i]}(\tau)w(s_i)-p\sum_i w(s_i)\right|\]
with:
\[w(s) = \exp\left(\alpha\frac{s-s_{\max}}{s_{\max}-s_{\min}}\right) \in [0, 1] \text{ when } \alpha>0\]
and similarly:
\[\tau_{top-p}\propto\exp\left(\frac{\epsilon U_{top-p}(\tau)}{2}\right)\]

\begin{figure}
\centering
\includegraphics[width=100mm,height=\textheight,keepaspectratio]{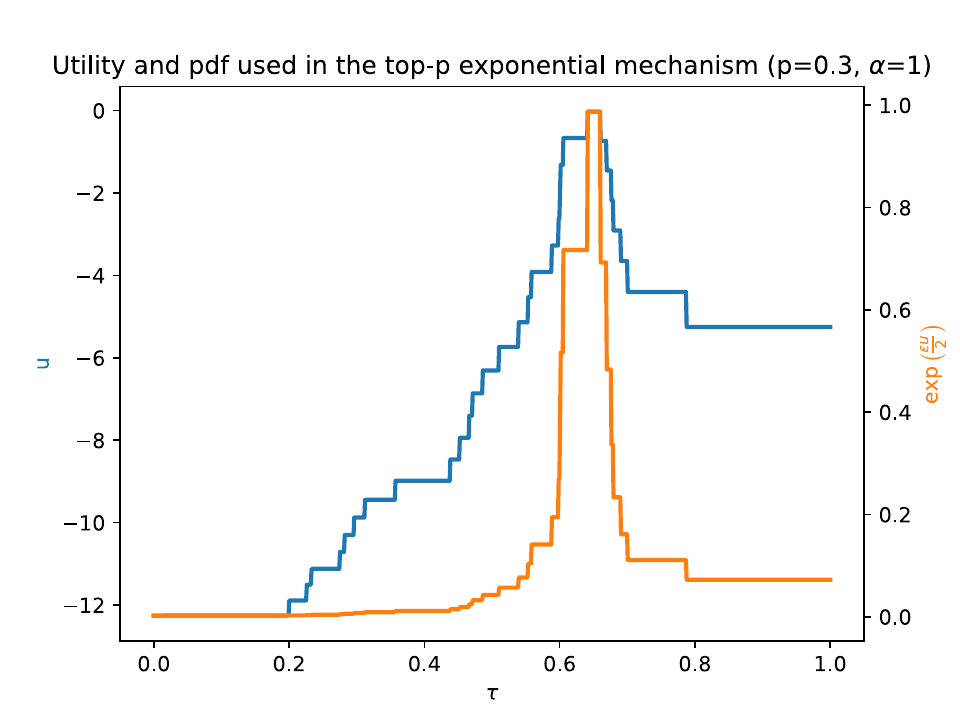}
\caption{The exponential mechanism for the top-p DP-threshold. For the
sake of clarity we chose a small number of documents: 30, and a large
\(\epsilon\): 1}\label{fig:toppexp}
\end{figure}

This utility function (see Fig.~\ref{fig:toppexp}) is parametrized by
\(\alpha\) which contrasts the differences in scores, and \(p\) which
select the share of \emph{total document weight} we want to select with
the mechanism.

Once the \(\tau_{top-p}\) threshold is sampled with DP, incurring a
small \emph{privacy loss}, it is safe to select the documents, the
similarity scores of which, are above it. They are then
\emph{aggregated} with DP in the DP ICL phase.

\subsection{Differentially Private In-Context
Learning}\label{differentially-private-in-context-learning}

In DP ICL, instead of sampling the next token from a query enhanced with
many documents:
\[L_{j+1}(\cdot) = \mathcal{L}\left(\cdot, \left<\left<q, d_{i_1}, d_{i_2}, \ldots d_{i_k}\right>_{RAG}, r_1, r_2,\ldots, r_j\right>\right)\]

we compute the distributions of the next token for many enhanced
queries, each of them with just one document: \[\left\{\begin{matrix}
L_{j+1,i_1}(\cdot) &=& \mathcal{L}\left(\cdot, \left<\left<q, d_{i_1}\right>_{RAG}, r_1, r_2,\ldots, r_j\right>\right)\\
L_{j+1,i_2}(\cdot) &=&\mathcal{L}\left(\cdot, \left<\left<q, d_{i_2}\right>_{RAG}, r_1, r_2,\ldots, r_j\right>\right)\\
\vdots\\
L_{j+1,i_k}(\cdot) &=&\mathcal{L}\left(\cdot, \left<\left<q, d_{i_k}\right>_{RAG}, r_1, r_2,\ldots, r_j\right>\right)
\end{matrix}\right.\]

We also compute the distribution of the next token, with some public
context:
\[L_{j+1, \text{pub}}(\cdot) = \mathcal{L}\left(\cdot, \left<\left<q, d_\text{pub}\right>_{RAG}, r_1, r_2,\ldots, r_j\right>\right)\]

Following (Tang et al. 2024), we sample a next token based on a DP
aggregation of the \(k+1\) distributions.

Contrary to (Tang et al. 2024) where they compare two mechanisms:
\emph{Gaussian} and \emph{Report Noisy Max}, and use a public prior with
\emph{Reduce Vocab Publicly} (RVP) we introduce a different mechanism:

\begin{itemize}
\tightlist
\item
  We use an exponential mechanism with a utility aggregating transformed
  log-probability vectors from all the enhanced queries.
\item
  We do not use Reduce Vocab Publicly (RVP), but a soft version,
  consisting in using the log-probabilities of a public response to
  boost or mute some tokens in a soft way.
\end{itemize}

We sample the next token from an exponential mechanism where the utility
is the aggregation of some function: \(l^\text{clipped}\) modulated by
the log-probabilities associated with the public query. The larger, the
\(\theta\), the closer the response will be to the one without the
private documents. This replaces RVP from (Tang et al. 2024).
\[U_{ICL}(r) = \theta \cdot \ln\left(L_{j+1, \text{pub}}(r)\right) + \sum_j l_{j+1,i_{j}}^\text{clipped}\left(r\right)\]

In the previous expression \(l^\text{clipped}\) is a clipped version of
\(l^\text{centered}\). \(l^\text{centered}\) is clipped to bound its
sensitivity (\(\infty\)-norm) in the exponential mechanism to some
\(C\).
\[l_{j+1,i_{j}}^\text{clipped}\left(r\right) = l_{j+1,i_{j}}^\text{centered}\left(r\right)\min\left(1, \frac{C}{\max_s \left|l_{j+1,i_{j}}^\text{centered}\left(s\right)\right|}\right)\]

Where \(l^\text{centered}\) is a centered version of \(l^\text{norm}\)
to minimize its \(\infty\)-norm without changing its impact in the
mechanism:
\[l_{j+1,i_{j}}^\text{centered}\left(r\right) = l_{j+1,i_{j}}^\text{norm}\left(r\right)-\frac{\max_sl_{j+1,i_{j}}^\text{norm}\left(s\right)+\min_sl_{j+1,i_{j}}^\text{norm}\left(s\right)}{2}\]

Where \(l^\text{norm}\) is a transformation of \(L\) putting more
emphasis on the large values of \(L\).
\[l_{j+1,i_{j}}^\text{norm}\left(r\right) = \frac{\exp\left[\alpha\cdot \left(\ln L_{j+1,i_{j}}\left(r\right) - \ln \max_s L_{j+1,i_{j}}\left(s\right)\right)\right]-1}{\alpha}\]

Indeed, for \(\alpha=1\) we simply compute a scaled and shifted version
of the probability:
\[l_{j+1,i_{j}}^\text{norm}\left(r\right) = \frac{L_{j+1,i_{j}}\left(r\right)}{\max_s L_{j+1,i_{j}}\left(s\right)}-1\]
for \(\alpha\) very small, we compute the log-probabilities:
\[l_{j+1,i_{j}}^\text{norm}\left(r\right) \approx \ln L_{j+1,i_{j}}\left(r\right) - \ln \max_s L_{j+1,i_{j}}\left(s\right)\]
and for \(\alpha\) very large, we get an indicator function:
\[l_{j+1,i_{j}}^\text{norm}\left(r\right) \approx 0 \text{ if } r=\text{argmax}_s L_{j+1,i_{j}}\left(s\right) \text{ and } -1 \text{ elsewhere}\]

After the utility is computed, the next token is sampled from:
\[r\propto \exp{\left(\frac{\epsilon U_{ICL}(r)}{2 C}\right)}\] In this
formula, the larger the \(\epsilon\) (privacy loss), or the smaller the
clipping \(C\) the closer we are to the most likely token.

The small (pure) privacy losses incurred during the retrieval phase and
the ICL phase are aggregated in a simple privacy accountant:
\href{https://github.com/google/differential-privacy/tree/main/python/dp_accounting}{github.com/google/differential-privacy/tree/main/python/dp\_accounting}.

The code of DP-RAG is available on
\href{https://github.com/sarus-tech/dp-rag}{github.com/sarus-tech/dp-rag}.

\section{Evaluation}\label{evaluation}

The DP-RAG algorithm, was tested on synthetic documents available on
Huggingface
\href{https://huggingface.co/datasets/sarus-tech/medical_dirichlet_phi3}{huggingface.co/datasets/sarus-tech/medical\_dirichlet\_phi3}.
The main benefit of using synthetic data is to make sure the LLM used
does not \emph{know} anything about the data a priori.

Here are a few examples documents (check the link above for more
examples):

\begin{longtable}[]{@{}
  >{\raggedright\arraybackslash}p{(\linewidth - 2\tabcolsep) * \real{0.5312}}
  >{\raggedright\arraybackslash}p{(\linewidth - 2\tabcolsep) * \real{0.4688}}@{}}
\toprule\noalign{}
\begin{minipage}[b]{\linewidth}\raggedright
Patient Question
\end{minipage} & \begin{minipage}[b]{\linewidth}\raggedright
Doctor Response
\end{minipage} \\
\midrule\noalign{}
\endhead
\bottomrule\noalign{}
\endlastfoot
I am Herman Schneider, and I'm grappling with a range of health issues.
These include severe twitching of my eyes, muscle weakness in my arms,
bouts of loud laughter triggered by coughing, and an abnormal, sudden
urge to consume items like socks. & Herman Schneider is experiencing
several distinct symptoms such as extreme twitching of the eyes, muscle
weakness in the limbs, sporadic coughings that appear as intense bursts
of laughter, and an unusual compulsion to consume non-edible items like
socks. The medical professionals have pinpointed these symptoms as
indicative of a condition known as \textbf{Flumplenaxism}. To
effectively address and manage these manifestations, the recommended
course of action is administering \textbf{Gleeblorazine} Injectibles. \\
& \\
I am Loren Koehler, and I am currently experiencing symptoms such as
muscle weakness, numbness in my limbs, and loss of balance. & Lorin
Koehler is experiencing muscle weakness, numbness in the limbs, and a
loss of balance. Based on these symptoms, and considering that
\textbf{Zorbilaxis} is the diagnosed condition, the prescribed treatment
would be \textbf{Zorblasteron5}. \\
& \\
Ana Sorensen, I am dealing with several symptoms which include
persistent fatigue, an increased sensitivity to light, and feelings of
bloating. & Ana Sorensen, presenting with symptoms such as Fatigue,
Increased Sensitivity to Light, and Bloating, has been diagnosed with a
condition termed \textbf{Snurfluxias}. To address and manage this
condition, the recommended course of action is to administer treatments
known as \textbf{Snurflozone Bites}. \\
\end{longtable}

When a query: \(q\), is submitted to DP-RAG, it is used in the Privacy
Unit Preserving Document Retrieval process, the retrieved documents are
fed into the DP-ICL phase of the algorithm. Here are a few examples of
interactions with the DP-RAG engine.

\begin{longtable}[]{@{}
  >{\raggedright\arraybackslash}p{(\linewidth - 2\tabcolsep) * \real{0.5312}}
  >{\raggedright\arraybackslash}p{(\linewidth - 2\tabcolsep) * \real{0.4688}}@{}}
\toprule\noalign{}
\begin{minipage}[b]{\linewidth}\raggedright
Question (\(q\))
\end{minipage} & \begin{minipage}[b]{\linewidth}\raggedright
DP-RAG Response (\(\epsilon \approx 5\))
\end{minipage} \\
\midrule\noalign{}
\endhead
\bottomrule\noalign{}
\endlastfoot
I am experiencing the following symptoms: Severe eye twitching, Muscle
weakness in the arms, Coughing fits of loud laughter, Sudden urge to eat
socks. What is my disease? & Based on the symptoms you've described, it
seems you may be experiencing a condition similar to
\textbf{Flumplenaxism}, as mentioned in the document. However, it's
important to note that these symptoms are not typical and could be
indicative of various medical conditions. \\
& \\
I am experiencing the following symptoms: Muscle weakness, Numbness in
limbs, Loss of balance. What is my disease? & Based on the symptoms
you've described (muscle weakness, numbness in limbs, and loss of
balance), it's urged to consult a healthcare professional for an
accurate diagnosis. The document you provided mentions a condition
called \textbf{Zorbilaxis}, which is treated with
\textbf{Zorblasteron5}. \\
& \\
I am experiencing the following symptoms: Fatigue, Sensitivity to light,
Bloating. What is my disease? & Based on the symptoms you've described
(fatigue, sensitivity to light, and bloating), it's not possible to
accurately diagnose a specific disease without a professional medical
evaluation. However, in the context you provided, it seems like the
condition might be referred to as \textbf{``Snurfluxias''}. \\
\end{longtable}

\begin{figure}
\centering
\pandocbounded{\includegraphics[keepaspectratio]{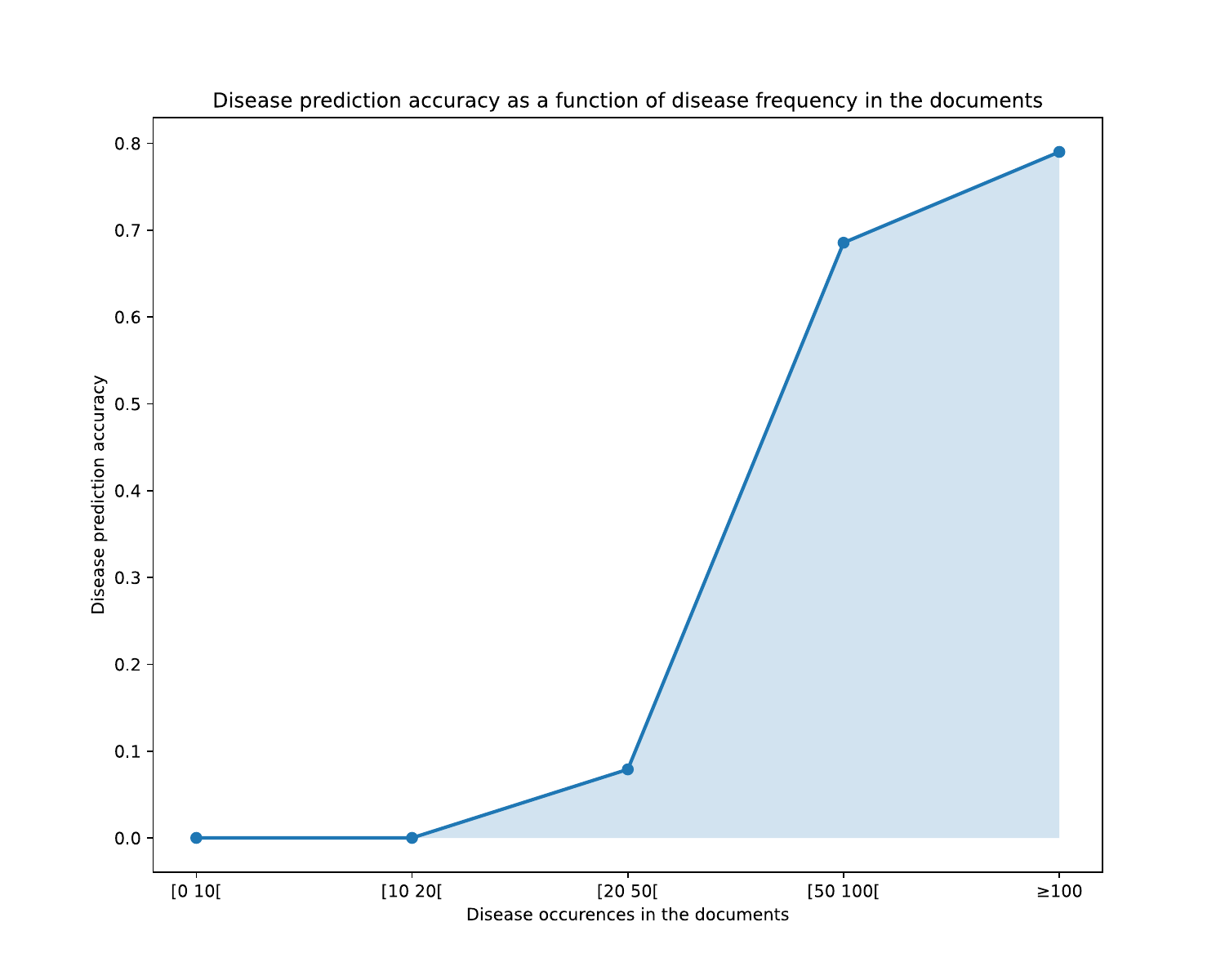}}
\caption{DP-RAG accuracy as a function of knowledge
specificity}\label{fig:accuracy}
\end{figure}

Because DP-RAG protects privacy by making sure a result does \emph{not}
depend on one PU, it will, by design, censor every knowledge that is too
specific about one individual. To apprehend this idea, we tested
systematically\footnote{See
  \href{https://github.com/sarus-tech/dp-rag/blob/main/results/evaluation.json}{github.com/sarus-tech/dp-rag/blob/main/results/evaluation.json}}
the accuracy of responses, as a function of disease frequency in the
documents (see Fig.~\ref{fig:accuracy}).

With around 5000 documents and privacy parameters set to
\(\epsilon \approx 5\), and, \(\delta=1e-3\), we got the result in
Fig.~\ref{fig:accuracy}.

The results show the accuracy becomes reasonable when at least 100
documents hold a similar piece of information.

\section{Conclusion}\label{conclusion}

Overall DP-RAG, provides a viable approach to private RAG in contexts
where documents are organized by individual (e.g.~Electronic Health
Records financial statements) and where sufficiently many documents
cover the question asked so that no one individual has an impact on the
response.

To improve the \emph{accuracy / privacy tradeoff}, one can:

\begin{itemize}
\tightlist
\item
  Ask question with shorter responses (and limit the number of tokens
  generated).
\item
  Make sure many documents are related to the question.
\item
  Increase the impact of the public prior (\(\theta\)) if some elements
  of the response are public.
\end{itemize}

\section{Bibliography}\label{refs}


\begin{CSLReferences}{1}{0}
\bibitem[\citeproctext]{ref-Abadi_2016}
Abadi, Martin, Andy Chu, Ian Goodfellow, H. Brendan McMahan, Ilya
Mironov, Kunal Talwar, and Li Zhang. 2016. {``Deep Learning with
Differential Privacy.''} In \emph{Proceedings of the 2016 ACM SIGSAC
Conference on Computer and Communications Security}. CCS'16. ACM.
\url{https://doi.org/10.1145/2976749.2978318}.

\bibitem[\citeproctext]{ref-bu2023differentially}
Bu, Zhiqi, Yu-Xiang Wang, Sheng Zha, and George Karypis. 2023.
{``Differentially Private Optimization on Large Model at Small Cost.''}
In \emph{International Conference on Machine Learning}, 3192--3218.
PMLR.

\bibitem[\citeproctext]{ref-carlini2021}
Carlini, Nicholas, Florian Tramèr, Eric Wallace, Matthew Jagielski,
Ariel Herbert-Voss, Katherine Lee, Adam Roberts, et al. 2021.
{``Extracting Training Data from Large Language Models.''} In \emph{30th
USENIX Security Symposium (USENIX Security 21)}, 2633--50. USENIX
Association.
\url{https://www.usenix.org/conference/usenixsecurity21/presentation/carlini-extracting}.

\bibitem[\citeproctext]{ref-dwork2014algorithmic}
Dwork, Cynthia, Aaron Roth, et al. 2014. {``The Algorithmic Foundations
of Differential Privacy.''} \emph{Foundations and
Trends{\textregistered} in Theoretical Computer Science} 9 (3--4):
211--407.

\bibitem[\citeproctext]{ref-hong2024dpoptmakelargelanguage}
Hong, Junyuan, Jiachen T. Wang, Chenhui Zhang, Zhangheng Li, Bo Li, and
Zhangyang Wang. 2024. {``DP-OPT: Make Large Language Model Your
Privacy-Preserving Prompt Engineer.''}
\url{https://arxiv.org/abs/2312.03724}.

\bibitem[\citeproctext]{ref-lebensold2024dprdmadaptingdiffusionmodels}
Lebensold, Jonathan, Maziar Sanjabi, Pietro Astolfi, Adriana
Romero-Soriano, Kamalika Chaudhuri, Mike Rabbat, and Chuan Guo. 2024.
{``DP-RDM: Adapting Diffusion Models to Private Domains Without
Fine-Tuning.''} \url{https://arxiv.org/abs/2403.14421}.

\bibitem[\citeproctext]{ref-lewis2021retrievalaugmentedgenerationknowledgeintensivenlp}
Lewis, Patrick, Ethan Perez, Aleksandra Piktus, Fabio Petroni, Vladimir
Karpukhin, Naman Goyal, Heinrich Küttler, et al. 2021.
{``Retrieval-Augmented Generation for Knowledge-Intensive NLP Tasks.''}
\url{https://arxiv.org/abs/2005.11401}.

\bibitem[\citeproctext]{ref-lin2024differentiallyprivatesyntheticdata}
Lin, Zinan, Sivakanth Gopi, Janardhan Kulkarni, Harsha Nori, and Sergey
Yekhanin. 2024. {``Differentially Private Synthetic Data via Foundation
Model APIs 1: Images.''} \url{https://arxiv.org/abs/2305.15560}.

\bibitem[\citeproctext]{ref-Ponomareva_2023}
Ponomareva, Natalia, Hussein Hazimeh, Alex Kurakin, Zheng Xu, Carson
Denison, H. Brendan McMahan, Sergei Vassilvitskii, Steve Chien, and
Abhradeep Guha Thakurta. 2023. {``How to DP-Fy ML: A Practical Guide to
Machine Learning with Differential Privacy.''} \emph{Journal of
Artificial Intelligence Research} 77 (July): 1113--1201.
\url{https://doi.org/10.1613/jair.1.14649}.

\bibitem[\citeproctext]{ref-qi2024followinstructionspillbeans}
Qi, Zhenting, Hanlin Zhang, Eric Xing, Sham Kakade, and Himabindu
Lakkaraju. 2024. {``Follow My Instruction and Spill the Beans: Scalable
Data Extraction from Retrieval-Augmented Generation Systems.''}
\url{https://arxiv.org/abs/2402.17840}.

\bibitem[\citeproctext]{ref-shokri2017}
Shokri, Reza, Marco Stronati, Congzheng Song, and Vitaly Shmatikov.
2017. {``Membership Inference Attacks Against Machine Learning
Models.''} In \emph{2017 IEEE Symposium on Security and Privacy (SP)},
3--18. \url{https://doi.org/10.1109/SP.2017.41}.

\bibitem[\citeproctext]{ref-sordoni2023jointpromptoptimizationstacked}
Sordoni, Alessandro, Xingdi Yuan, Marc-Alexandre Côté, Matheus Pereira,
Adam Trischler, Ziang Xiao, Arian Hosseini, Friederike Niedtner, and
Nicolas Le Roux. 2023. {``Joint Prompt Optimization of Stacked LLMs
Using Variational Inference.''} \url{https://arxiv.org/abs/2306.12509}.

\bibitem[\citeproctext]{ref-tang2024privacypreservingincontextlearningdifferentially}
Tang, Xinyu, Richard Shin, Huseyin A. Inan, Andre Manoel, Fatemehsadat
Mireshghallah, Zinan Lin, Sivakanth Gopi, Janardhan Kulkarni, and Robert
Sim. 2024. {``Privacy-Preserving in-Context Learning with Differentially
Private Few-Shot Generation.''} \url{https://arxiv.org/abs/2309.11765}.

\bibitem[\citeproctext]{ref-wu2023privacypreservingincontextlearninglarge}
Wu, Tong, Ashwinee Panda, Jiachen T. Wang, and Prateek Mittal. 2023.
{``Privacy-Preserving in-Context Learning for Large Language Models.''}
\url{https://arxiv.org/abs/2305.01639}.

\bibitem[\citeproctext]{ref-xie2024differentiallyprivatesyntheticdata}
Xie, Chulin, Zinan Lin, Arturs Backurs, Sivakanth Gopi, Da Yu, Huseyin A
Inan, Harsha Nori, et al. 2024. {``Differentially Private Synthetic Data
via Foundation Model APIs 2: Text.''}
\url{https://arxiv.org/abs/2403.01749}.

\bibitem[\citeproctext]{ref-yue2023synthetictextgenerationdifferential}
Yue, Xiang, Huseyin A. Inan, Xuechen Li, Girish Kumar, Julia McAnallen,
Hoda Shajari, Huan Sun, David Levitan, and Robert Sim. 2023.
{``Synthetic Text Generation with Differential Privacy: A Simple and
Practical Recipe.''} \url{https://arxiv.org/abs/2210.14348}.

\bibitem[\citeproctext]{ref-zeng2024mitigatingprivacyissuesretrievalaugmented}
Zeng, Shenglai, Jiankun Zhang, Pengfei He, Jie Ren, Tianqi Zheng,
Hanqing Lu, Han Xu, Hui Liu, Yue Xing, and Jiliang Tang. 2024.
{``Mitigating the Privacy Issues in Retrieval-Augmented Generation (RAG)
via Pure Synthetic Data.''} \url{https://arxiv.org/abs/2406.14773}.

\bibitem[\citeproctext]{ref-zhou2023largelanguagemodelshumanlevel}
Zhou, Yongchao, Andrei Ioan Muresanu, Ziwen Han, Keiran Paster, Silviu
Pitis, Harris Chan, and Jimmy Ba. 2023. {``Large Language Models Are
Human-Level Prompt Engineers.''} \url{https://arxiv.org/abs/2211.01910}.

\end{CSLReferences}
\end{document}